\documentclass[sigconf]{acmart}
\usepackage{tabulary}

\AtBeginDocument{%
  \providecommand\BibTeX{{%
    \normalfont B\kern-0.5em{\scshape i\kern-0.25em b}\kern-0.8em\TeX}}}

\setcopyright{acmcopyright}
\copyrightyear{2023}
\acmYear{2023}
\acmDOI{XXXXXXX.XXXXXXX}

\acmConference[WebScience '23]{WebScience}{April 30--May 01,
  2023}{Austin, TX}
\acmPrice{15.00}
\acmISBN{978-1-4503-XXXX-X/18/06}

\usepackage{color}

\begin{document}

\title{Anger Breeds Controversy: Analyzing Controversy and Emotions on Reddit}

\author{Kai Chen}
\affiliation{%
  \institution{USC Information Sciences Institute}
  \streetaddress{4676 Admiralty Way}
  \city{Marina del Rey}
  \state{CA}
  \country{USA}}
\email{kchen035@usc.edu}

\author{Zihao He}
\affiliation{%
  \institution{USC Information Sciences Institute}
  \streetaddress{4676 Admiralty Way}
  \city{Marina del Rey}
  \state{CA}
  \country{USA}}
\email{zihaoh@usc.edu}

\author{Rong-Ching Chang}
\affiliation{%
  \institution{University of California, Davis}
  \city{Davis}
  \state{CA}
  \country{USA}
}
\email{rocchang@ucdavis.edu}

\author{Jonathan May}
\affiliation{%
  \institution{USC Information Sciences Institute}
  \streetaddress{4676 Admiralty Way}
  \city{Marina del Rey}
  \state{CA}
  \country{USA}}
 \email{jonmay@isi.edu}

\author{Kristina Lerman}
\affiliation{%
  \institution{USC Information Sciences Institute}
  \streetaddress{4676 Admiralty Way}
  \city{Marina del Rey}
  \state{CA}
  \country{USA}}
  \email{lerman@isi.edu}

\renewcommand{\shortauthors}{Chen and He, et al.}

\begin{abstract}
Emotions play an important role in interpersonal interactions and social conflict, yet their function in the development of controversy and disagreement in online conversations has not been explored. To address this gap, we study controversy on Reddit, a popular network of online discussion forums. We collect discussions from a wide variety of topical forums and use emotion detection to recognize a range of emotions from text, including anger, fear, joy, admiration, etc. Our study has three main findings. First, controversial comments express more anger and less admiration, joy and optimism than non-controversial comments. Second, controversial comments affect emotions of downstream comments in a discussion, usually resulting in long-term increase in anger and a decrease in positive emotions, although the magnitude and direction of emotional change depends on the forum. Finally, we show that emotions help better predict which comments will become controversial. Understanding emotional dynamics of online discussions can help communities to better manage conversations.

\end{abstract}





\maketitle


\section{Introduction}
The social web has linked millions of people worldwide,  creating ``digital town squares'' for exchanging ideas, opinions, and beliefs. On platforms like Reddit and Twitter, among many others, people post messages or respond to the messages posted by others. The low barriers to entry into global online conversations offers society many benefits, such as democratizing the production and distribution of information, reducing the power of traditional gatekeepers to decide what information gets attention, creating better ways for people to learn from the diverse experiences of others, and catalyzing mass protest movements. 
Unfortunately, the same mechanisms that lead to societal benefits are also responsible for creating unique new vulnerabilities. Exchanging diverse viewpoints within global online communities invites disagreement, which malicious actors and anti-social trolls exploit to derail conversations, spread misinformation, and inflame polarization. The rise in anti-social online behaviors has had profound consequences on society, undermining collective trust in institutions and in democracy itself~\cite{haidt2022atlantic}. 

Online communities have tried to reduce harmful speech by mediating discussions to remove messages that violate community norms due to toxicity, harassment, or personal attacks \cite{park-etal-2021-detecting-community}. However, manual moderation does not scale to the volume and speed of online conversations. Although machine moderation has improved in recent years, with tools that automatically recognize harassment, hate speech, and other types of toxic speech~\cite{macavaney2019hate, poletto2021resources, plaza2021comparing}, these methods treat the symptoms, rather than causes of the problem. In order to better identify and mediate controversy, we need to understand how controversy develops and derails conversations in open online communities before we can  effectively---and automatically---moderate them.



Researchers have attempted to identify controversial discussions in online communities using network approaches~\cite{10.1145/3140565} or features derived from user activity~\cite{10.1145/3447535.3462481}. Others have trained models to learn language cues associated with controversial comments~\cite{zayats-ostendorf-2018-conversation,park-etal-2021-detecting-community}. 
With advances in natural language processing, we are now able to move beyond these works to explore psycholinguistic dimensions of controversy. 
Specifically, we study how controversy and disagreement develop within online discussions through the lens of emotions.
We focus on emotions because they are the cornerstone of interpersonal interactions~\cite{van2016social} and shape the social response to conflict~\cite{bar2007collective}. 
Emotions are also important in online interactions and have been shown to contribute to the viral spread of topics~\cite{brady2021social,coviello2014detecting,bi2022emotions}. However, the role of emotions in the development of controversy or disagreement in online discussions has not been explored. 

We study Reddit, a popular network of online communities. Within Reddit's many topical forums, or `subreddits,' members post new topics for discussion, and others comment on these submissions or respond to the comments of others. Community members can upvote or downvote any comment, expressing their agreement or disagreement with it. Reddit automatically flags a comment as \emph{controversial}  if it has a large and similar number of upvotes and downvotes.  

To study how emotions affect the development of controversy and disagreement in online interactions, we pose the following research questions:
\begin{description}
\item[RQ1] Are controversial comments more emotional than non-controversial comments?
\item[RQ2] How do controversial comments change emotions in the discussion?
\item[RQ3] Can we identify controversial comments at time of creation, i.e., before they become controversial?
\end{description}

To answer these research questions, we use a state-of-the-art emotion detection method \cite{alhuzali2021spanemo} to measure a range of emotions expressed in text. We find that controversial comments on Reddit express substantially more anger and less joy, love and optimism than non-controversial comments. Although controversial comments represent a small fraction of all comments in a discussion---typically, only 3\% of comments are controversial---we show that discussions with at least one controversial comment also express more anger and less positive emotions like joy and love. To explain this observation, we investigate how controversial comments change emotions of the subsequent discussion by comparing the emotions expressed in comments that follow a controversial comment to the emotions expressed in the comments preceding it. We find that controversial comments set the long-term emotional tone of discussions. Finally, we show that adding emotions as features to a state-of-the-art controversial comment classification method leads to significant performance improvement. This enables us to predict whether a comment will become controversial, potentially allowing a moderator to step in to keep the conversation from becoming overheated.

We argue that, besides focusing on simple metrics like the number of upvotes and downvotes of comments, public media watchdogs and social media platforms should pay attention to the emotional tone of online discussions. Understanding the emotional dynamics of online conversations could help communities engage in more constructive dialog and prevent disagreement and controversy from derailing conversations.


\section{Related Work}
In this section, we briefly introduce controversy detection on different social media platforms, emotion detection, and a few recent works that have laid the groundwork by combining emotions and controversy detection. 

\subsection{Controversy Detection on Social Platforms}
Previous work has explored different methods to detect and predict controversy in online platforms. 
Some works leverage the network structures between the users, submissions, and text features for detection of controversial comments. 
For example, \citet{10.1145/3140565} construct conversation graphs on Twitter and characterize controversy based on graph structures, such as random walks, betweenness centrality, and low dimensional embeddings. They define a measure of controversy based on random walks, which measures how likely a random user joining a controversial discussion is to be exposed to the dominant authority of each side in the debate. They show that this method using graph structural features is better at identifying controversial topics than those using content-based features. 
\citet{zayats-ostendorf-2018-conversation} train a graph-structured bidirectional LSTM to predict the popularity of comments in Reddit discussions and further use language cues to help identify controversial comments.
Similarly, \citet{park-etal-2021-detecting-community} detect norm violations on Reddit, leveraging LSTM and pretrained language models, which are more suitable for sequential data.

\citet{10.1145/3447535.3462481} identify controversial comments in multilingual discussions on Reddit by training a linear classifier on features of comments and discussions. They explore different types of features, including lexical features of the comments themselves, the predecessors, and successors. 
They find that user activities (such as the rate at which people comment before and after a controversial comment, and the number of preceding comments) produce the most discriminating features. 
One confounding factor is that discussions with controversial comments receive more attention since they are flagged by Reddit, making them easier to find through its user interface. The increased attention affects the evolution of controversial discussions.
\citet{jang2018explaining} summarize controversy through stance-indicativeness, articulation, and topic relevance and the evaluation shows that their summaries based on these lexicon features has a better understanding of the controversy.

Besides detecting individual controversial comments, some works have focused on identifying controversial submissions, which initiate discussions on Reddit.
\citet{hessel-lee-2019-somethings} and \citet{zhong-etal-2020-integrating} define controversial submissions by ratio of upvotes to all votes. 
\citet{hessel-lee-2019-somethings} focus on both the textual contents of comments as well as the discussion hierarchy. They conclude that the textual contents are significantly more helpful for detecting submission controversy but fail to generalize to different forums (subreddits); however, structural features are more generalizable.

\subsection{Emotion Recognition}
To understand emotions in human text, early research uses dictionary-based methods to measure the sentiment expressed in messages by counting positive or negative words they contain~\cite{golder2011diurnal,bollen2011twitter,chen2014building,mejova2014controversy,dori2013detecting}. 
Another popular approach measures emotions in text along the dimensions of valence and arousal, with the former capturing the level of pleasure, or positive sentiment, expressed in text, and the latter capturing the level of activation induced by the emotion.
This approach relies on lexicons, e.g., the WKB lexicon~\cite{warriner2013norms}, that include valence and arousal scores of common English words. After lemmatizing the input text, they average the scores of terms that match the lexicon features. Using these methods, researchers find that the sentiment of tweets display characteristic diurnal and weekly patterns of mood variation~\cite{golder2011diurnal} and are able to track the geographic distribution of emotional wellbeing~\cite{jaidka2020estimating}. 

These lexicon-based approaches, however, do not account for context and are difficult to extend to multilingual data due to the effort required to label words. To address these challenges, a new generation of methods based on large language models enables a wider range of emotional expressions to be quantified at scale~\cite{alhuzali2021spanemo}. 
These methods benefit from the availability of large-scale datasets of sentences that have associated emotion labels. For example, \citet{mohammad2018semeval} provide a corpus of tweets annotated with emotion labels in English, Arabic and Spanish. Multilingual transformers like XLM-T \cite{xlm2021barbieri, wolf2020transformers} have been trained on this data, extending emotion detection capability to multilingual settings.
In addition, GoEmotions \cite{demszky2020goemotions} is a dataset of 58k English Reddit comments with up to 28 different categories of emotion.

\subsection{Emotions and Controversy Online}
Research exploring the role of emotions in controversy on a large scale is largely under-explored. \citet{bi2022emotions} studies the role of emotions in the diffusion of posts on Facebook. She finds evidence that both positive and negative emotions are directly associated with the diffusion of highly controversial topics. \citet{mejova2014controversy} categorize news into controversial and non-controversial and find that controversial news tends to use more negative emotional words. Similarly, \citet{stieglitz2013emotions} and \citet{brady2017emotion} discover that emotional tweets tend to spread wider and faster based on analysis of political discussions on Twitter. 

Building upon prior works, we focus on psycholinguistic indicators, specifically emotions expressed in comments. We further extend the contribution by studying the relationship and dynamics between emotions and controversial comments in different granularity, covering individual comments, how controversial comments impact the succeeding comments in discussions, and the ability to detect controversial comments using emotional cues. 

\section{DATA}

Reddit is  a popular social platform for user discussions. It consists of a wide range of topical forums, or subreddits. In each subreddit, a user can start a discussion by posting a new submission, which other users can comment on or respond to the comments of others.
We use Pushshift API~\cite{baumgartner2020pushshift} to collect Reddit discussions. Pushshift has archives of Reddit data dating back to 2005, which includes complete discussions and metadata, such as the controversial tag. A comment is automatically flagged by Reddit as controversial when the numbers of upvotes and downvotes it has are both high and very similar. We further define a controversial discussion as the one that includes at least one controversial comment. 

From the collected discussions, we create two datasets -- Dataset I: Popular Forums (used in Sec. \ref{sec:emo-contro-comments} and Sec. \ref{sec:contro-change}) and Dataset II: Mutilingual Forums (used in Sec. \ref{sec:contro-pred}).

\subsection{Dataset I: Popular Forums}
\label{sec:generalization-data}
We collect data from the 100 most popular subreddits on Reddit based on the number of subscribers.
For subreddits that were very large, we randomly under-sample discussions so that the number of discussions from all subreddits were roughly similar. 
We then filter out discussions with fewer than five comments, and discard ten subreddits that disallow users from commenting after 2022, such as \emph{r/announcement}.
The remaining 90 subreddits cover a large variety of topics, such as art (\textit{r/Art}, \textit{r/pics}), music (\textit{r/Music}, \textit{r/listentothis}), sports (\textit{r/sports}, \textit{r/nba}), politics (\textit{r/politics}, \textit{r/news}), science (\textit{r/science}, \textit{r/space}),  humor (\textit{r/jokes}, \textit{r/Animalsbeingbros}, \textit{r/facepalm}), gender (\textit{r/TwoXChromosomes}), advice (\textit{r/lifehacks}, \textit{r/LifeProTips}), gaming (\textit{r/PS4}, \textit{r/Minecraft}), and  emotional reactions (\textit{r/aww}, \textit{r/wholesomememes}, \textit{r/mademesmile}), among many others. The complete list of the 90 subreddits are shown on the y-axis of Figure \ref{fig:emotion-generalization-all}.
We call this dataset of popular forums Dataset I and show the statistics in Table~\ref{tab:popular_stats}. 



\begin{table}[ht]
    \centering
    \begin{tabular}{l|cccc}
\hline
                                                                             & \textbf{min} & \textbf{max} & \textbf{mean} & \textbf{median} \\ \hline
\begin{tabular}[c]{@{}l@{}}avg. \# of comments\\ per discussion\end{tabular} & 7.9          & 557.6        & 104.4         & 82              \\ \hline
\begin{tabular}[c]{@{}l@{}}ratio of controversial \\ comments\end{tabular}   & 0.3\%        & 9.7\%        & 3\%           & 2.8\%           \\ \hline
\begin{tabular}[c]{@{}l@{}}ratio of moderated\\ comments\end{tabular}        & 3.1\%        & 57.9\%       & 12.7\%        & 9.5\%           \\ \hline
\end{tabular}
    \caption{Statistics of discussions in Dataset I: Popular Forums.}
    \label{tab:popular_stats}
\end{table}


\subsection{Dataset II: Multilingual Forums}
\label{sec:prediction_data}
For controversy detection, we sample six subreddits from the aforementioned popular forums covering four different categories: science (\emph{r/science} and \emph{r/technology}), question \& answer (\emph{r/AskScience} and \emph{r/AskReddit}), news (\emph{r/news} and \emph{r/worldnews}).  To further demonstrate our approach's generalizability to discussions in languages other than English, we add multilingual discussions from subreddits \emph{r/france} (in French) and \emph{r/de} (in German). 
We call this dataset of multilingual forums Dataset II and report the statistics in Table \ref{tab:prediction-dataset-statistics}.
%

\begin{table*}[ht]
\centering
\begin{tabular}{@{}l@{}|llllllll@{}}
\hline
& r/science & r/technology & r/news  & r/worldnews & r/AskReddit & r/AskScience & r/france & r/de    \\ \hline
number of discussions           & 32,744     & 102,246       & 8,691    & 17,858       & 188,177      & 73,665        & 50,558    & 63,058   \\
number of comments              & 1,681,039   & 1,482,271      & 2,116,989 & 2,693,907     & 6,615,721     & 1,681,039      & 1,628,475  & 1,845,356 \\
average discussion length       & 51.3      & 14.4         & 243.5   & 150.8       & 35.2        & 5.7          & 32.2     & 29.2    \\
ratio of controversial comments & 4.4\%    & 5.8\%       & 7.4\%   & 7.9\%       & 1.2\%      & 1.1\%       & 5.9\%   & 5.1\%   \\
ratio of removed comments       & 42.5\%   & 10.8\%      & 18.5\% & 11.2\%     & 5.8\%      & 47.3\%      & 6.3\%   & 6.3\%  \\ \hline
\end{tabular}
\caption{Statistics of Dataset II: Multilingual Forums.}
\label{tab:prediction-dataset-statistics}
\end{table*}


\section{Methods and Results}
We answer our research questions by analyzing emotions and controversy of online discussions.
We define a discussion to be controversial if it has at least one comment that has been tagged as controversial by Reddit \footnote{Results do not differ qualitatively when using a higher threshold of the number of controversial comments to define controversial discussions}. As a reminder, a comment is tagged as ``controversial'' if it has the same (or similar) number of upvotes as downvotes, and both numbers are large.

\subsection{Overview of Emotions on Reddit}
To recognize emotions expressed in text, we use a multilingual emotion detection model from \citet{chochlakis2022using,chochlakis2022leveraging}. The model is based on SpanEmo~\cite{alhuzali2021spanemo} that is the state-of-the-art in emotion detection. 
The original backbone language model BERT \cite{devlin-etal-2019-bert} is replaced with multiligual XLM-T \cite{xlm2021barbieri, wolf2020transformers}, which is more suitable in a multilingual setting to handle text inputs in large number of languages.
The model was finetuned on SemEval 2018 Task 1 E-c data \cite{mohammad2018semeval} and GoEmotions \cite{demszky2020goemotions}, with ten simplified \textit{emotion clusters}: ``Anger, Hate, Contempt, Disgust,'' ``Embarrassment, Guilt, Shame, Sadness,'' ``Admiration, Love,'' ``Optimism, Hope,'' ``Joy, Happiness,'' ``Pride, National Pride,'' ``Fear, Pessimism,'' ``Amusement,'' ``Other Positive Emotions,'' and ``Other Negative Emotions.''
Further details of the model and performance are described in 
\cite{chochlakis2022using, chochlakis2022leveraging}.\footnote{https://github.com/gchochla/Demux-MEmo}
Given input text, the model returns one scalar value per emotion, indicating the \emph{confidence} that the emotion is present. Since it is a multi-label classification setting, the model can assign multiple (or no) emotions to text input. 
Therefore, for each input Reddit comment, we have a 10d confidence vector of the ten emotion clusters.

\begin{figure*}[ht]
    \centering
    \begin{tabular}{c}
    \includegraphics[width=\linewidth]{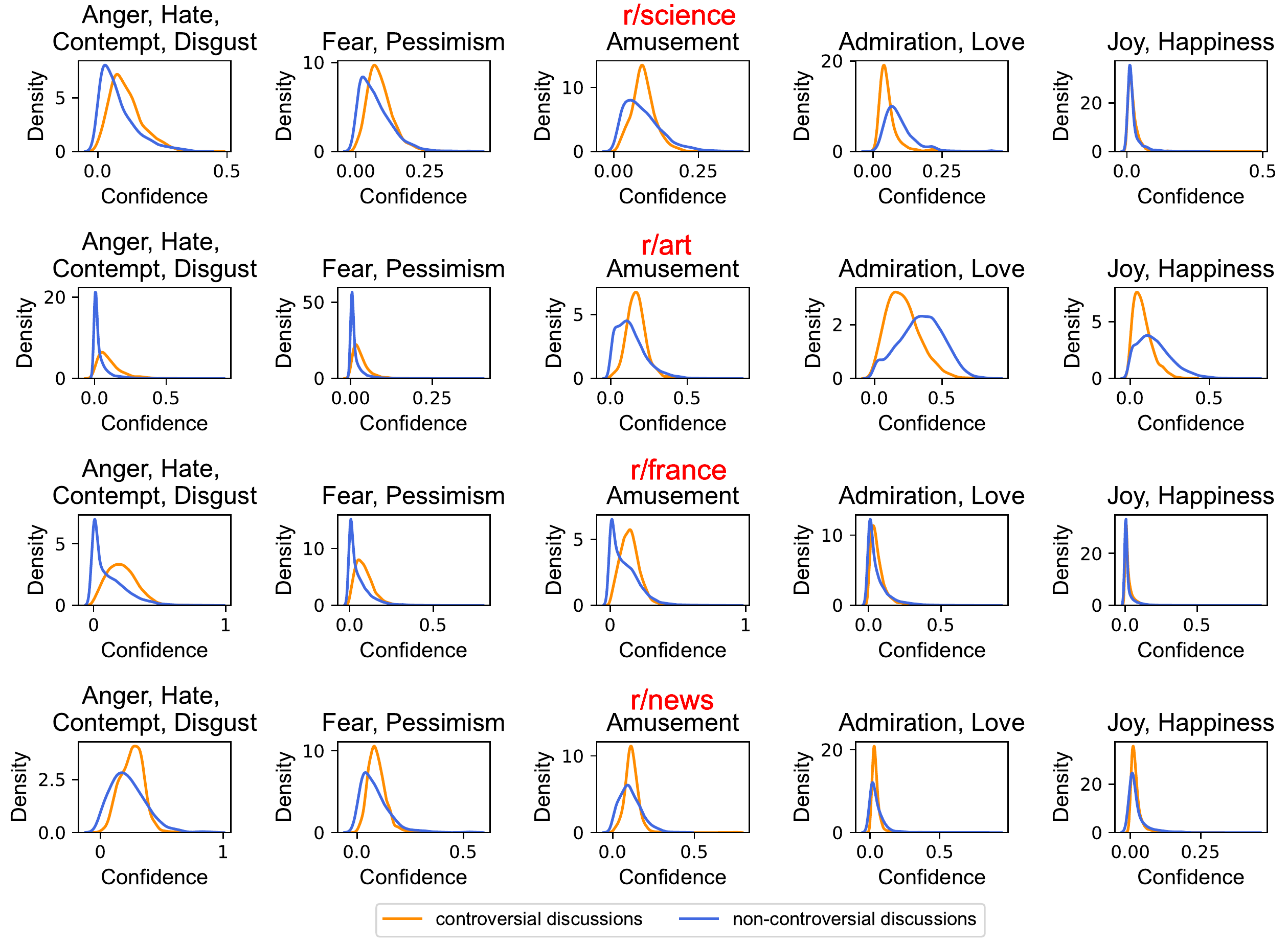}
    \end{tabular}
    \caption{Distributions of five emotion clusters of dicussions on four subreddits. The emotion confidence values of each discussion are averaged from those of each comment within it.}
    \label{fig:emotion-generalization-comment}
\end{figure*}

Consider a discussion of length $L$ (i.e. it has $L$ comments). We measure the confidence $f_i(e)$ of emotion $e$ expressed in comment $i$ of the discussion using the emotion detection model. By averaging over all $L$ comments in a discussion, we obtain $\hat{f}(e)$, its average emotion confidence.
Figure~\ref{fig:emotion-generalization-comment} shows the distribution of $\hat{f}(e)$ for five emotions of controversial discussions (orange curve) and non-controversial discussions (blue curve) on four subreddits.
We observe that the controversial and non-controversial discussions have largely different distributions on some emotions.
Anger/\-Hate/\-Contempt/\-Disgust, for example, has systematically higher confidence in controversial discussions on \textit{r/france} and \textit{r/news} compared to non-controversial discussions. On the other hand, the confidence of the Admiration/\-Love emotion is higher in non-controversial discussions on \textit{r/art} and \textit{r/science}. Our research quantifies these differences and helps explain how they arise.

\subsection{RQ1: Emotions in Controversial Comments}
\label{sec:emo-contro-comments}




To answer our first research question, we compare emotions expressed in controversial comments 
to emotions in non-controversial comments. 
Let ${CC}$ be the set of \textbf{c}ontroversial \textbf{c}omments in a subreddit, and $NC$ be the set of \textbf{n}on-controversial \textbf{c}omments in the same subreddit. 
We define emotion gap $\delta_e$  as the difference between the mean confidence of emotion $e$ in controversial comments and its mean confidence  in  non-controversial comments:
$$
\delta_e = \frac{1}{|CC|} \sum_{i \in CC} f_i (e) - \frac{1}{|NC|}\sum_{i \in NC} f_i(e).
$$
We calculate  $\delta_e$ separately for each emotion and subreddit.

\begin{figure}[ht]
    \centering
    \includegraphics[width=\linewidth]{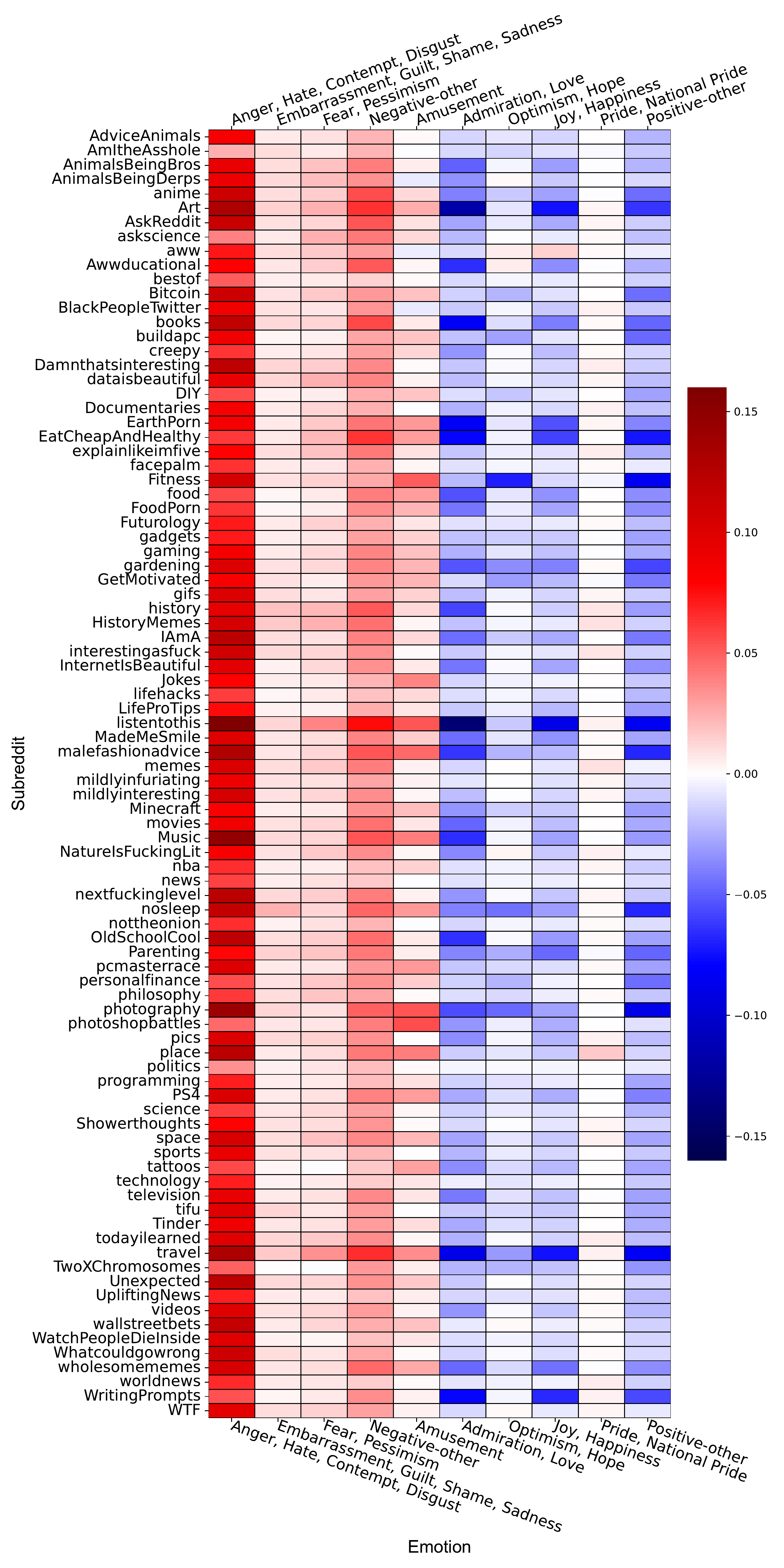}
    \caption{Emotion gaps between controversial and non-controversial comments for all subreddits in Dataset I. Red indicates higher emotion confidence for controversial comments, blue indicates higher emotion confidence for non-controversial comments, and white indicates equal emotion confidence between them. Color saturation denotes the magnitude of emotion gap.
    }
    \label{fig:emotion-generalization-all}
\end{figure}

Figure~\ref{fig:emotion-generalization-all} shows the emotion gaps $\delta_e$ for all ten emotion clusters and across all 90 subreddits in Dataset I.
We observe strong global trends. The Anger/\-Hate/\-Disgust emotion cluster is consistently stronger in controversial comments than in non-controversial comments (as indicated by bright red colors in Fig.~\ref{fig:emotion-generalization-all}). Negative other emotions are also higher in controversial comments, but there are no strong differences in other negative emotions like Embarrassment/\-Guilt/\-Shame and Fear/\-Pessimism.  
In contrast, positive emotions are stronger in non-controversial comments than in controversial comments. 
For example, Admiration/\-Love, Joy/\-Happiness, and Positive-other are all much less common in controversial comments than in non-controversial comments (as indicated by bright blue color in Fig.~\ref{fig:emotion-generalization-all}). 
The emotion Amusement appears to be stronger in controversial comments in half of the forums, but rarely weaker. This emotion is usually used to denote text that is funny, but sometimes also captures sarcasm. This suggests that controversial comments are often funny or sarcastic, though not in all forums.

There are many differences across subreddits in the strength of the emotion gap. For example, compared to other forums, subreddits \textit{r/Music}, \textit{r/listentothis}, \textit{r/Art} have strongest differences across all emotions. Controversial comments on these forums have much more Anger than non-controversial comments, but also much less Admiration and Joy than non-controversial comments. 
Surprisingly, the forums that we expected to have more controversy, like \textit{r/politics} and \textit{r/AmItheAsshole}, show smaller emotional differences between controversial and non-controversial comments.

These results answer our first research question. \textit{Controversial comments are angrier than non-controversial comments and express less positive emotion, like love, joy and optimism}. 
These differences also apply to controversial discussions, though the magnitude of the emotional gap is reduced:  
compared to non-controversial discussions, controversial discussions express more anger and less admiration and joy. 
Controversial comments alone do not explain the difference in the emotionality of controversial discussions, since they represent a small share of all comments (see Table~\ref{tab:popular_stats}). Instead, controversial comments change the emotional tone of subsequent comments, which we explore next.

\subsection{RQ2: Controversial Comments Change Emotions in Discussions}
\label{sec:contro-change}
In this section, we explore how controversial comments shape emotions in the discussions. To quantify the impact  of a controversial comment $i$ on emotion $e$ in a discussion, we calculate the difference between the average confidence of $e$ in comments posted \textit{after} the comment $i$ and comments that came \textit{before} it.
Let $L$ be the length of a discussion, and $i$ the position of a controversial comment within the discussion. We refer to comments in positions $\{1, \ldots, i-1\}$ as \textit{predecessors} of the controversial comment, and comments in positions  $\{i+1, \ldots, L\}$ as \textit{successors} of comment $i$. The emotional impact $\theta^i_e$ of the controversial comment $i$ is:
$$
\theta^i_e=\frac{1}{(L-i)} \sum_{j > i} f_j (e) - \frac{1}{(i-1)}\sum_{j <i} f_j(e).
$$
If $\theta^i_e>0$, the controversial comment $i$ leads to more emotion $e$ in subsequent comments; otherwise, if $\theta^i_e<0$, the comment $i$ reduces that emotion in the discussion.



\begin{figure}[h!]
    \centering
    \includegraphics[width=0.97\linewidth]{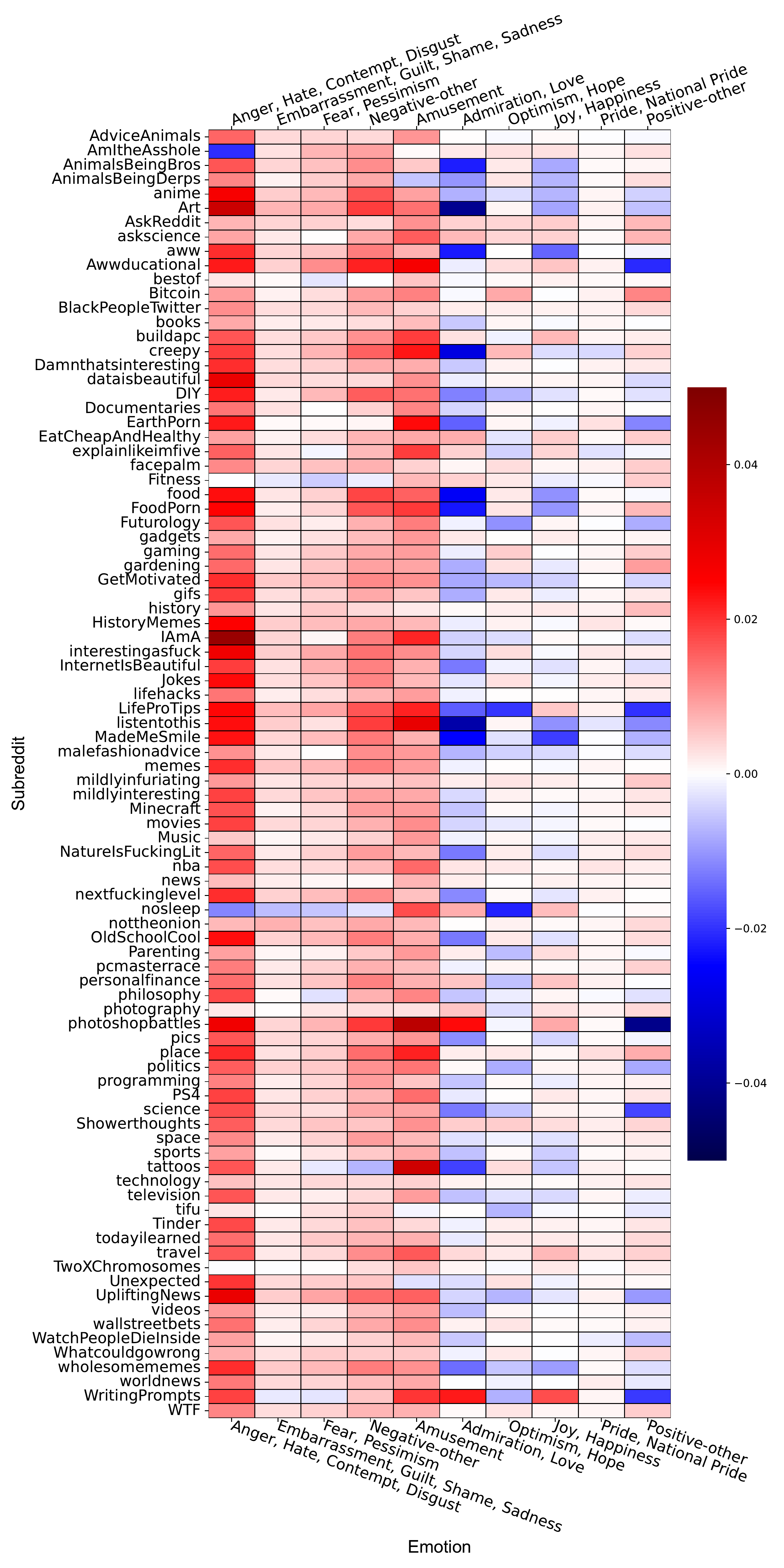}
    \caption{
    Emotional impact of controversial comments. Cells show the difference between the average emotion confidence of successors of a controversial comment and its predecessors in Dataset I.
    Red indicates a long-term increase in emotions following controversial comments, blue indicates long-term decrease in emotions, and white indicates no change in emotions following controversial comments. 
    Color saturation denotes the magnitude of the impact.
    }
    \label{fig:emotion-before-after-generalization}
\end{figure}

To measure the overall impact of controversial comments on emotions, we average the impact of all controversial comments within each subreddit's discussions. 
Figure~\ref{fig:emotion-before-after-generalization} shows this quantity  across all emotions and all 90 subreddits. Most subreddits follow the same pattern: negative emotions like Anger/\-Hate/\-Contempt/\-Disgust, Fear/\-Pessimism and other negative emotions rise after a controversial comment and positive emotions generally, though not always, fall. There are some exceptions to this trend. In  \emph{r/nosleep} and \emph{r/AmItheAsshole} subreddits, anger decreases after a controversial comment. 
Moreover, positive emotions, such as Admiration/\-Love and Joy/\-Happiness, in \emph{r/photoshopbattle} and \emph{r/WritingPrompts} increase after a controversial comment. 
Amusement rises systematically in almost all subreddits, on par with negative emotions. This suggests rising sarcasm in controversial discussions.

These findings answer our second research question and suggest that \textit{controversial comments lead to long-term changes in the emotional tone of discussions,} typically, not only raising anger of downstream comments but also reducing positive emotions of discussions. However, different communities respond differently to controversy, resulting in some deviations from this pattern. Therefore, automatic tools to moderate controversial comments will need to take the varying community context into account.

\begin{figure*}
    \centering
    \begin{tabular}{c}
    \includegraphics[width=\linewidth]{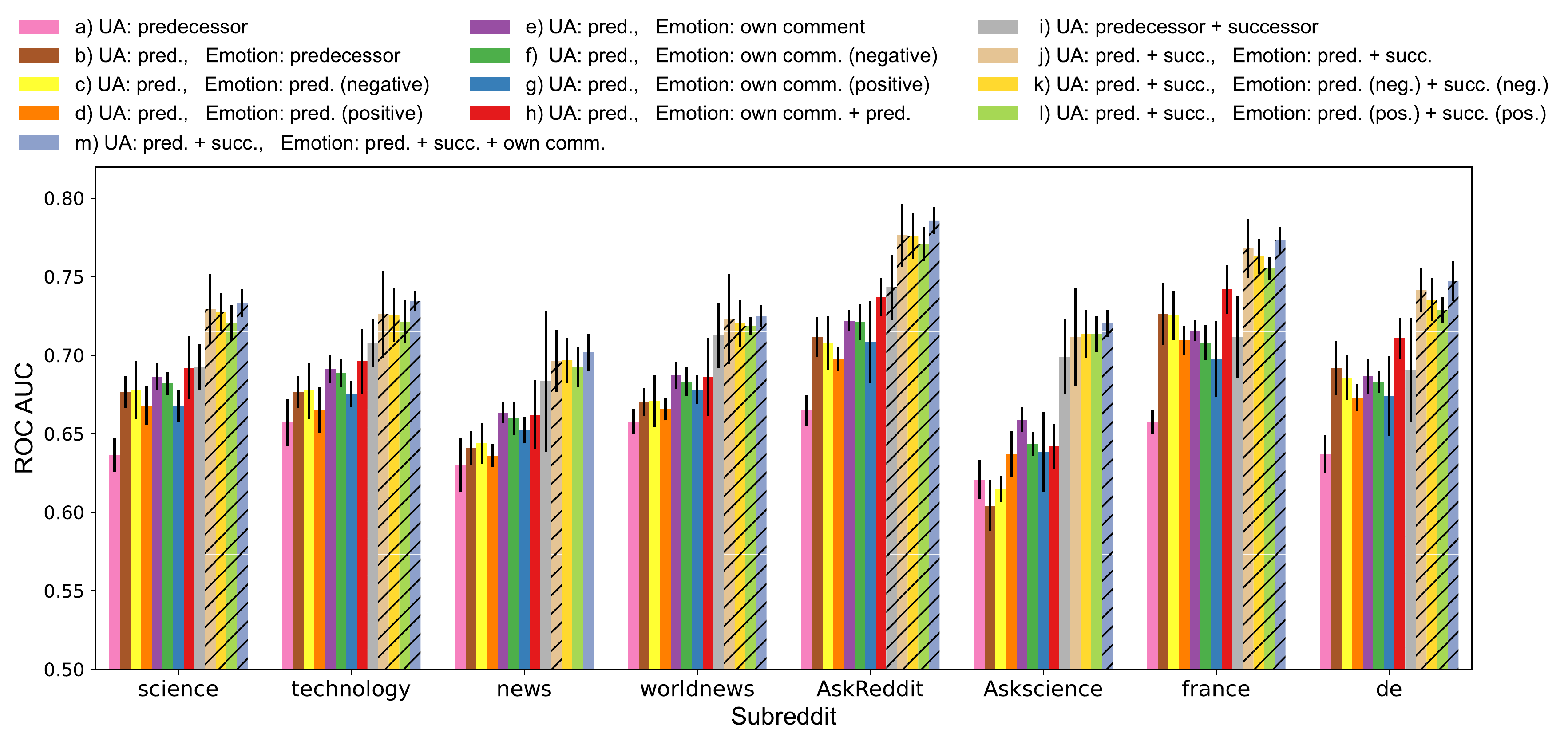}
    \end{tabular}
    \caption{Performance (ROC AUC) of controversial comment prediction using ten-fold cross validation on eight subreddits using thirteen features sets, including user activity (UA) features and emotion features of the current comment, its predecessors, and its successors.
    Results using successor features are shadowed. Overall the models that take in emotion features achieve a better performance.
We use user activity (UA) features as baseline and test different combinations of UA and emotion features. All comments are in English except for the \emph{r/france} (in French) and \emph{r/de} (in German) subreddits.}
    \label{fig:prediction_result}
\end{figure*}

\subsection{RQ3: Predicting Controversy from Emotions}
\label{sec:contro-pred}

\citet{10.1145/3447535.3462481} address the task of predicting whether a comment $i$ in a discussion on Reddit is controversial using a variety of lexical and user activity features. They find that the most predictive features were ones related to user activity, which they calculate based on comments that preceded the comment $i$ in a discussion (i.e., predecessors) and comments that succeeded it (i.e., successors): 
\begin{itemize}
  \item Predecessor features: number of comments preceding the current comment $i$, number of unique authors of preceding comments, elapsed time from the first predecessor till the current comment, and the average elapsed time between predecessors.
  \item Successor features: number of comments posted after the comment $i$, number of unique authors of succeeding comments, eplased time from the current comment to the last successor, and average time between successors.
\end{itemize}
We supplement these features with emotions, and use the emotions expressed in the current comment, and those expressed in predecessors or successors, as features.
\begin{itemize}
  \item All emotions: confidence of emotion clusters in predecessor comments, successor comments, or the comment $i$'s own emotion.
  \item Positive emotions: subset of emotions that includes Admi\-ra\-tion/\-Love,  Optimism/\-Hope, Joy/\-Happiness, Pride, other positive emotions.
  \item Negative emotions: subset of emotions that includes Anger/\-Hate\-/\-Contempt\-/\-Disgust, Embarrassment/\-Sadness, Fear/\-Pe\-ssi\-mism, and other negative emotions.
\end{itemize}

To better understand the impact of emotions on predicting controversial comments, we construct different feature sets to be used by the classification model (as shown in Fig. \ref{fig:prediction_result}). 
Following \citet{10.1145/3447535.3462481}, we used Gradient Boosted Decision Trees with ten-fold cross-validation for the prediction task.\footnote{We also tried using a pretrained language model \cite{devlin-etal-2019-bert} taking in the original comments as inputs without feature engineering for predicting comment controversy, but the model failed to converge.} 
We conducted a grid search to select hyperparameters that achieve the best performance on the validation set. 

We applied the model to predict whether a comment is controversial using Dataset II (Sec.~\ref{sec:prediction_data}). This data includes discussions from six popular subreddits (in English) and also discussions in French and German, demonstrating the utility of our approach to multilingual settings. 
The data is highly imbalanced in that controversial comments make up only 5\% of all comments; therefore, we under-sample non-controversial comments to create a balanced dataset for testing.
We use ROC-AUC to measure classification performance.






Figure~\ref{fig:prediction_result} shows the results of the controversial comment classification separately for the eight subreddits. Overall, adding emotions to the model consistently improves performance 
compared to using only activity features. 
Such non-trivial improvement demonstrates the effectiveness of emotions in controversial comment detection.

The best performance is achieved on  \emph{r/AskReddit}, with AUC of 78.6\%, followed by \textit{r/france} and \textit{r/de} with AUC of 77.3\% and 74.7\%, respectively. These scores represent  4.2\%, 6.2\%, and 5.7\%, respectively,  improvement in performance compared to using  activity features only (comparing feature set m and feature set i). 
Interestingly, these three subreddits are also the  least moderated forums in our data (see Table~\ref{tab:prediction-dataset-statistics}), with the lowest ratio of comments removed by moderators. This suggests that it is easier to identify controversial comments in less-moderated or unmoderated forums.

In addition to the overall comparison between different feature sets and different subreddits, there are some intriguing observations on feature selection.
First, the model that uses comment's own emotion as a feature (feature sets e, f, g) outperforms models that use emotions of predecessor comments (feature sets b, c, d). 
Taking a deeper look, we observe that the average  length of discussions on \emph{r/news} and \emph{r/worldnews} is extremely high (on average 243.5 comments per discussion). As a result, the emotion features of predecessors are too noisy due to many non-controversial comments among the predecessors.
Second, negative emotions are more effective than positive emotions in predicting controversy, suggesting that controversy is better conveyed by negative emotions. 
Features of successors (feature sets i, j, k, l, m) are more helpful in strongly moderated subreddits, such as in \emph{r/askscience}, because these features still leverage the structure information from removed comments. 

Finally, although using features of predecessors and the current comment does not perform as well as using all features (including successor features), it represents a more realistic prediction scenario. On this task, adding emotions to the feature set significantly improves classification performance across all subreddits. Our results show emotions help identify controversial comments \textbf{before} they become controversial.

\section{Discussion and Conclusion}

Emotions, or feelings, are fundamental to human experience, and play a critical role in  the formation of beliefs, social interactions~\cite{van2016social}, and interpersonal conflict~\cite{bar2007collective}. Our study demonstrates that emotions also shape the  evolution of controversial discussions in online communities. Leveraging emotional cues present in language, we identify a range of positive and negative emotions expressed in the text of comments. Our large-scale study of discussions on more than 90 subreddits shows that controversial comments have stronger negative emotions, especially anger, and fewer positive emotions than non-controversial comments. Although controversial comments represent a small share of all comments in a discussion, they shift the emotional tone of the entire discussion, leading to angrier and less positive subsequent comments. 
We also show that an emotionally aware classification model could better recognize comments that will become controversial, even in multilingual discussions.


Our work suggests that moderating controversial comments may help improve the emotional tone of a discussion. It is possible to catch such comments at the time of their creation, and step in to help the author regulate the negative emotions such comments express.  Reducing the long-term impact of controversial comments on the discussion will help improve the overall quality of the discussions.

\bibliographystyle{ACM-Reference-Format}
\bibliography{references}

\end{document}